\documentclass[letterpaper,twocolumn,fleqn]{article}

\usepackage{ist}
\usepackage{graphicx}
\usepackage{amsmath,amssymb}
\usepackage{cite}
\usepackage[table]{xcolor}
\usepackage{multirow}
\usepackage{adjustbox}
\usepackage{hyperref}
\usepackage{placeins}
\hypersetup{colorlinks=true, linkcolor=black, citecolor=black, urlcolor=black}

\makeatletter
\long\def\@makecaption#1#2{%
  \ifx\@captype\@tablestring
    {\small\sffamily \textbf{#1.}\hspace{0.5em}#2}\@tablecaptionsepspace
  \else
    \@figurecaptionsepspace
    \setbox\@tempboxa\hbox{\scriptsize\sffamily\itshape {\bfseries #1.}~~ #2}%
    \ifdim \wd\@tempboxa >\hsize
      \setbox\@tempboxa\hbox{\scriptsize\sffamily\itshape {\bfseries #1.}~~ }%
      \parbox[t]{\hsize}{\scriptsize\sffamily\itshape \noindent\unhbox\@tempboxa#2}%
    \else
      \hbox to\hsize{\scriptsize\sffamily\itshape\box\@tempboxa\hfil}%
    \fi
  \fi}
\makeatother

\newcommand{\figref}[1]{Fig.~\ref{#1}}
\newcommand{\tabref}[1]{Table~\ref{#1}}

\title{Benchmarking color judgment in multimodal LLMs with CIE-computable ground truth}

\author{
Shuwei Yue$^{1,}$, Kaida Xiao$^{2}$, Zhihua Wang$^{3}$, and Qiao Chen$^{1}$\newline
$^{1}$Shenzhen Polytechnic University, Shenzhen, China\newline
$^{2}$University of Leeds, Leeds, United Kingdom\newline
$^{3}$Sun Yat-sen University, Guangzhou, China\newline
$^{}$Corresponding author: shuwei.yue@connect.polyu.hk
}

\date{}
\begin{document}
\maketitle
\thispagestyle{empty}
\begin{abstract}
Multimodal large language models (MLLMs) are now used as
zero-shot judges of image quality, but their color judgments are
rarely tested with colorimetric ground truth. We present a
five-task benchmark based on the CIE colorimetric framework: AWB
pair judgment, skin-tone bias direction, CCT estimation,
$\Delta E_{2000}$ pair ordering, and color naming under changing
illuminants. We test eight MLLM endpoints, including four
Qwen3-VL variants, GLM-4.5V, GPT-5.4, Claude-Opus-4.7, and
Gemini-3.1-Pro. The models often solve coarse binary AWB
judgments, but they remain weak on numeric CCT estimation and on
some fine color-difference judgments. On ColorChecker-24 stimuli,
the three closed-weight frontier models reach CCT MAE of
715--1056\,K, compared with 1358--2133\,K for the five
open-weight models. A non-ColorChecker control shows that this
advantage is partly template-dependent. The results suggest that
current MLLMs have useful color words and coarse visual priors,
but their color judgments are not reliably calibrated to
CIE-computable color differences.
\end{abstract}

\section{Introduction}
\label{sec:intro}

Color is a core concern of any imaging pipeline. Cameras, displays,
and printers all spend large compute budgets on white balance, color
correction, and tone mapping, and all of these are tuned against
physical color measurements. Multimodal large language models
(MLLMs) are now being used in this pipeline as zero-shot
quality judges~\cite{huang2025qdoc, li2026qprobe},
but few benchmarks isolate color judgment against
colorimetrically computable ground truth.

Recent MLLM-as-judge work matches human raters on aesthetic and
document-quality tasks~\cite{huang2025qdoc, li2026qprobe}. However, these benchmarks use human-annotated
ground truth, which mixes the model's output with rater preference. This is a real drawback for color, where a $\Delta E$ around one
is often treated as a just-noticeable difference under controlled
viewing conditions --- a precision that human raters cannot
reliably deliver.

\begin{figure}[t]
\centering
\includegraphics[width=\linewidth]{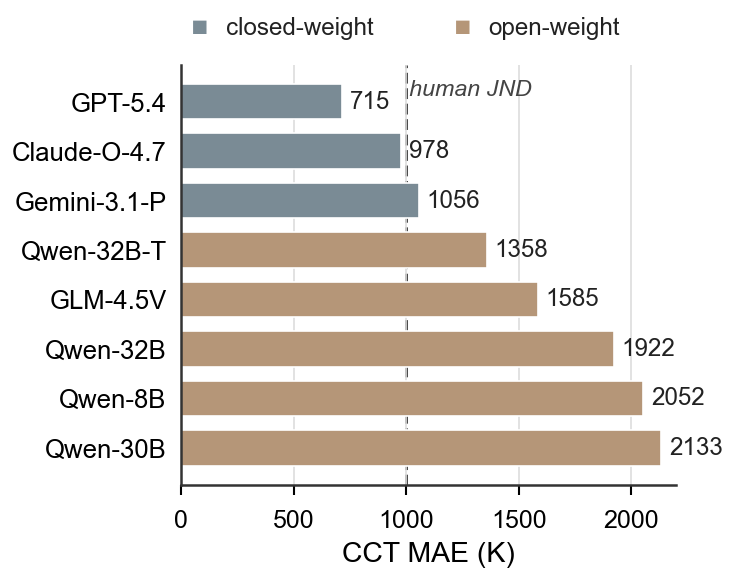}
\caption{\textbf{Headline CCT result.} Closed-weight frontier
endpoints reach lower CCT MAE on ColorChecker-24 than the tested
open-weight endpoints. The dashed 1000\,K line is a visual
reference only; Kelvin is not perceptually
uniform~\cite{pearce2014}.}
\label{fig:teaser}
\end{figure}

Color is one of the few visual attributes with a computable
ground truth. Unlike object recognition or counting, color has
standardized chromatic-adaptation transforms, correlated color
temperature (CCT) loci, CIELab, and $\Delta E_{2000}$ formulas,
which together let us quantify errors in colorimetric units
without human annotation. This is the property our benchmark uses.

We build on two recent lines of work. The MLLM-IQA
work~\cite{huang2025qdoc, li2026qprobe,
li2025lowlevelvlm} shows that MLLMs can be prompted as visual judges,
but its stimuli mix many low-level cues (sharpness, noise, exposure)
and do not isolate color as a single variable. The generative-color
work~\cite{fang2026toovivid, tsai2025colormecorrectly} shows that even flagship T2I models miss
named colors, but it tests \emph{generation}, not \emph{perception}.
Our benchmark targets the perception side and uses the standard
ColorChecker-24 stimulus design from classical color-constancy
research~\cite{gehler2008shi, ciurea2003sfu}.
\medskip

\noindent\textbf{Contributions}~In this paper, we (1) introduce a
five-task color benchmark whose ground truth is computed from the
CIE colorimetric framework~\cite{cie2018}, not from human
annotation; (2) evaluate eight MLLM endpoints across four vendors
and show that coarse AWB judgments, numeric CCT estimation,
perceptual $\Delta E_{2000}$ ordering, and grounded color naming
fail in different ways; and (3) separate the observed failures
into endpoint-family effects (warm-bias and $\Delta E$ ordering
plateau within Qwen), a daylight-band CCT central-tendency prior
shared by the open-weight cluster, a D65-grounded naming
bottleneck on Task E, and a model-specific pick-vs-name asymmetry
in Claude-Opus-4.7. The benchmark, raw responses, and analysis
scripts are released.

\section{Related work}
\label{sec:related}

Recent MLLM-as-judge benchmarks evaluate aesthetic, document, and
low-level image quality with human labels~\cite{huang2025qdoc,
li2026qprobe, li2025lowlevelvlm}, but their
stimuli mix sharpness, exposure, noise, layout, and semantics, and
do not isolate color against computable ground truth. A parallel
line on generative color fidelity reports that text-to-image
models often miss named colors or over-saturate
them~\cite{fang2026toovivid, tsai2025colormecorrectly}, but it
tests generation rather than perception. Classical color-constancy
benchmarks such as Gehler--Shi and SFU
automatic white balance (AWB)~\cite{gehler2008shi, ciurea2003sfu} provide the
ColorChecker stimulus convention and illuminant-estimation context,
to which recent learning-based methods add pure-color and
cross-sensor priors~\cite{yue2024crosssensor}. We reuse that
stimulus convention to evaluate prompt-based MLLMs rather than
specialised AWB networks. Language-only color-cognition
work~\cite{abdou2021lmcolor} motivates our interpretation of
vocabulary-driven naming failures.

\section{Method}
\label{sec:method}

\begin{figure*}[!t]
\centering
\includegraphics[width=\linewidth]{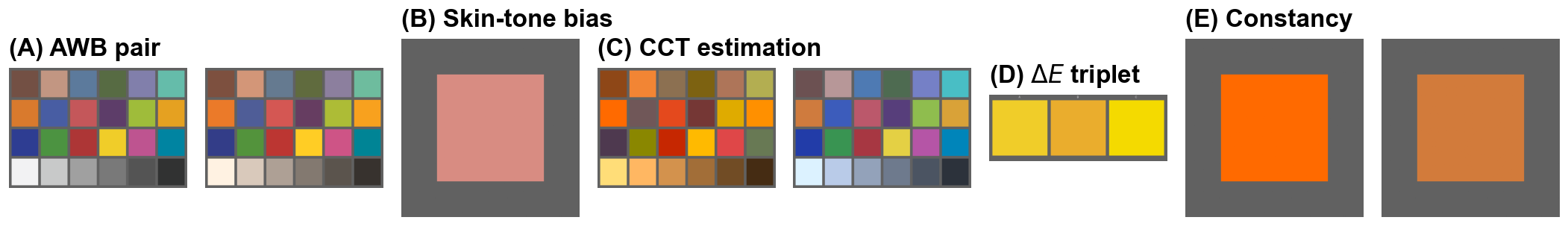}
\caption{\textbf{Representative stimuli for the five
tasks.}
(A)~Task A AWB pair (correct vs.\ wrong).
(B)~Task B skin-tone bias direction.
(C)~Task C ColorChecker rendered at 2700 and 10000\,K.
(D)~Task D $\Delta E$ ordering triplet.
(E)~Task E color constancy --- the same patch under 2700 and
9000\,K. All stimuli are 480$\times$320 PNGs.}
\label{fig:stimuli}
\end{figure*}

The benchmark contains five tasks that test different parts of
MLLM color judgment: pair-wise discrimination (Tasks A and D),
direction naming (Task B), numeric regression (Task C), and
cross-illuminant naming consistency (Task E). All stimuli are
generated with CIE colorimetric formulas. Tasks A--D have
colorimetric ground truth, and Task E uses a fixed
Lab-nearest-neighbor color-name reference. All prompts return
single-line JSON, so the benchmark can run on any
OpenAI-compatible endpoint without model-specific code.
\figref{fig:stimuli} shows one representative stimulus per task.

\noindent\textbf{Task A --- AWB pair judgment}~The model sees two
ColorChecker24 charts in random order, differing only in white
balance: one adapted to D65, the other multiplied by a fixed
R/B/G/M cast gain in linear sRGB. The model picks the
correctly-balanced image and names the cast direction. 24 pairs:
4 cast types $\times$ 6 CCTs. Prompts are zero-shot with a fixed
JSON schema; the cast label refers to the incorrect image (full
prompts in the released repo).

\noindent\textbf{Task B --- skin-tone bias}~A single dark-skin or
light-skin ColorChecker patch is shifted in CIELab by a known
$(\Delta a^{*}, \Delta b^{*})$ offset along one of 8 hue-clock
directions, at three sizes ($\Delta E\!\approx\!4, 8, 14$). 48
stimuli. The model returns direction and size.

\noindent\textbf{Task C --- CCT estimation}~The chart is rendered
as if captured under a target white point at CCT
$T \in \{2700, 3500, 4200, 5000, 5700, 6500, 7500, 8500, 10000\}$\,K
with no white-balance correction. The white point is taken from the
Planckian locus for $T \le 3500$\,K and from the CIE D-illuminant
locus for $T \ge 4200$\,K (the CIE D polynomial is undefined below
4000\,K). The model returns an integer CCT in Kelvin. 27 stimuli
(9 CCTs $\times$ 3 noise seeds).

\noindent\textbf{Task D --- $\Delta E_{2000}$ pair ordering}~Three
patches A, B, C are shown side by side. The model picks which pair,
$(A,B)$ or $(A,C)$, has a larger color difference. Ground-truth
$\Delta E_{2000}$ is computed from the underlying Lab values using
the reference implementation~\cite{sharma2005ciede2000}; pair gaps
are kept at $\geq 4$ units to avoid forced ties. 24 stimuli.

\noindent\textbf{Task E --- color naming under illuminant}~The
same underlying ColorChecker patch value is rendered with no
chromatic adaptation under $T \in \{2700, 4200, 6500, 7500,
9000\}$\,K. The model names the color from the 11 basic-color
terms of Berlin \& Kay (red, orange, yellow, green, blue, purple,
pink, brown, black, gray, white). The D65 reference name is
assigned by $\Delta E_{ab}$ nearest-neighbour to ISCC-NBS
centroids in CIELab; there is no human annotation in the
evaluation loop. 40 stimuli (8 patches $\times$ 5 illuminants).
The raw constancy index is the per-patch fraction of renderings
that get the same dominant name; FM4 reports the conditional
variant.
\\

\noindent\textbf{Stimulus generation}~All stimuli follow four
design rules. First, every correct answer can be computed from CIE
015:2018 colorimetric formulas~\cite{cie2018}, so no annotator is
needed. Second, each stimulus varies along a single color dimension
while luminance, layout, and image content are held fixed. Third,
prompts are zero-shot with a single fixed template per task,
temperature\,$=0$, and no in-context examples, so reported scores
reflect the model's prior rather than prompt engineering. Fourth,
stimuli are 480$\times$320 PNGs and prompts return single-line
JSON, so the benchmark runs against any OpenAI-compatible endpoint
with no changes.

The 24 ColorChecker patches are stored as sRGB-encoded values
under D65 (a colorimetric shortcut --- they are not full
reflectance spectra; v2 will use measured BabelColor CC spectra)
and decoded to linear sRGB. The scene white at the target CCT is
taken from the Planckian locus for $T \le 3500$\,K (where the
CIE D polynomial is undefined) and from the CIE D locus above. A
Bradford CAT then maps patches from D65 to the scene white;
$\sigma{=}0.01$ Gaussian gain noise is added; values are clipped
from below at zero (no upper clip), re-encoded to sRGB, and
rasterised as a $480\times320$ PNG.
\\

\noindent\textbf{Models and protocol}~We evaluate eight MLLMs from
four vendors. The Qwen3-VL family gives the primary axis: four
variants on a single training stack isolate \emph{parameter scale}
and \emph{reasoning mode} --- Qwen3-VL-8B-Instruct (small dense),
30B-A3B-Instruct (Mixture-of-Experts, MoE; 3B active), 32B-Instruct (dense flagship),
and 32B-Thinking (chain-of-thought). GLM-4.5V (Zhipu, 106B) is a
cross-family probe trained on an independent corpus and visual
encoder. The April 2026 closed-weight frontier models --- GPT-5.4
(OpenAI), Claude-Opus-4.7 (Anthropic), and
Gemini-3.1-Pro-preview (Google) --- serve as a contrastive
control. A failure that survives the Qwen$\to$GLM probe but
breaks here points to a shared-prior origin rather than a
single-stack artifact. Qwen and GLM are accessed via SiliconFlow; the three
frontier models via OpenRouter (model IDs
\texttt{openai/gpt-5.4}, \texttt{anthropic/claude-opus-4.7},
\texttt{google/gemini-3.1-pro-preview}). All calls use
temperature\,$=0$ and max\_tokens$=512$; the larger cap
accommodates the Thinking variant's chain-of-thought. Because
closed-weight endpoints can change without notice, all raw
responses (including the call timestamps) are released with the
benchmark.

\section{Experiments and results}
\label{sec:results}

We report headline accuracy on each of the five tasks
(\tabref{tab:headline}), then examine five repeated failure patterns
(FM1--FM5) followed by a non-ColorChecker robustness check.
Statistical intervals throughout are 1000-resample bootstrap 95\%
CIs unless noted otherwise.

\noindent\textbf{Headline accuracy}~\tabref{tab:headline} gives
the per-(model, task) headline metric. 
Five repeated failure patterns emerge from this matrix --- each
isolated to a particular vendor cluster, model family, or task
axis. We describe FM1--FM5 , then report a non-CC
robustness check that tests how much of the headline split is
template-dependent.
\medskip

\begin{table*}[t]
\centering
\caption{Headline accuracy on the five color tasks for eight MLLMs,
grouped by access mode. Each cell: mean with 95\% bootstrap CI
(1000 resamples). \textbf{Bold} = best, \underline{underline} =
2nd-best per column. Chance: A which = 0.50, A bias = 0.25
(4-way), B dir$\pm$1 = 0.375 (3 of 8 clock positions), D order =
0.50. ``1.00 [.86,1]'' uses the Clopper-Pearson 95\% lower bound
for 24/24 successes (vs.\ the zero-variance bootstrap [1,1]).
Task E ``raw constancy'' rewards stable-but-wrong names; read
together with the conditional analysis (FM4).}
\label{tab:headline}
\renewcommand{\arraystretch}{1.25}
\resizebox{\textwidth}{!}{%
\begin{tabular}{l|ccccccc}
\hline
\textbf{Model}
  & \textbf{A which}                   & \textbf{A bias}
  & \textbf{B dir$\pm$1}               & \textbf{C MAE\,K}
  & \textbf{C $\le$500K}               & \textbf{D order}
  & \textbf{E raw const.} \\
  & {\scriptsize $n{=}24,\ \uparrow$}  & {\scriptsize $n{=}24,\ \uparrow$}
  & {\scriptsize $n{=}48,\ \uparrow$}  & {\scriptsize $n{=}27,\ \downarrow$}
  & {\scriptsize $n{=}27,\ \uparrow$}  & {\scriptsize $n{=}24,\ \uparrow$}
  & {\scriptsize $n{=}8$ patch, $\uparrow$} \\
\hline
\multicolumn{8}{l}{\emph{Qwen3-VL family --- open-weight, scale + reasoning axis}} \\
\quad 8B-Instruct       & 0.58 [.38,.79] & 0.25 [.08,.46] & 0.60 [.48,.73] & 2052 [1522,2537] & 0.11 [.00,.26] & 0.79 [.62,.92] & 0.78 [.65,.90] \\
\quad 30B-A3B-Instruct  & 0.67 [.46,.83] & 0.25 [.08,.46] & 0.52 [.39,.67] & 2133 [1511,2774] & 0.15 [.04,.30] & 0.79 [.62,.92] & 0.68 [.55,.80] \\
\quad 32B-Instruct      & \textbf{1.00 [.86,1]}    & \textbf{1.00 [.86,1]}    & 0.60 [.48,.75] & 1922 [1459,2411] & 0.22 [.07,.41] & 0.79 [.62,.92] & \underline{0.85 [.73,.95]} \\
\quad 32B-Thinking      & \textbf{1.00 [.86,1]}    & \textbf{1.00 [.86,1]}    & 0.67 [.54,.81] & 1358 [\phantom{0}950,1808] & 0.33 [.17,.54] & 0.75 [.58,.92] & \underline{0.85 [.73,.95]} \\
\hline
\multicolumn{8}{l}{\emph{GLM-4.5V --- open-weight cross-family probe (independent stack)}} \\
\quad GLM-4.5V          & \textbf{1.00 [.86,1]}    & 0.92 [.79,1.0] & 0.60 [.46,.75] & 1585 [1115,2048] & 0.30 [.15,.48] & \underline{0.88 [.71,1.0]} & 0.78 [.65,.90] \\
\hline
\multicolumn{8}{l}{\emph{Closed-weight frontier (April 2026)}} \\
\quad GPT-5.4           & \textbf{1.00 [.86,1]}    & \textbf{1.00 [.86,1]}    & \textbf{0.79 [.67,.90]} & \textbf{715 [\phantom{0}563,\phantom{0}863]} & 0.30 [.11,.48] & \textbf{0.92 [.79,1.0]} & 0.80 [.68,.90] \\
\quad Claude-Opus-4.7   & 0.67 [.50,.88] & \textbf{1.00 [.86,1]}    & 0.56 [.44,.71] & \underline{978 [\phantom{0}589,1411]} & \textbf{0.56 [.37,.74]} & \underline{0.88 [.75,1.0]} & 0.83 [.70,.95] \\
\quad Gemini-3.1-Pro    & \textbf{1.00 [.86,1]}    & \textbf{1.00 [.86,1]}    & \underline{0.75 [.63,.88]} & 1056 [\phantom{0}722,1418] & 0.33 [.15,.52] & \underline{0.88 [.75,1.0]} & \underline{0.85 [.73,.95]} \\
\hline
\end{tabular}
}
\end{table*}

\noindent\textbf{Failure mode 1: warm-bias collapse
(Qwen-specific)}~Qwen-8B and Qwen-30B-MoE both collapse to the
``warm'' label and achieve chance-level A-bias accuracy (0.25),
while Qwen-32B, GLM-4.5V, and the three frontier models reach
$0.92$--$1.00$. Because the two failing variants are also the two
with the fewest active parameters per token (Qwen-30B-MoE has only
3B active), the benchmark cannot separate a Qwen-family prior
from a low-active-parameter effect; the cleanest reading is the
conjunction. Warm-color vocabulary is over-represented in
language-corpus statistics~\cite{abdou2021lmcolor}, consistent
with a vocabulary-driven rather than perceptual origin for the
collapse.
\\

\begin{figure*}[!t]
\centering
\includegraphics[width=\linewidth]{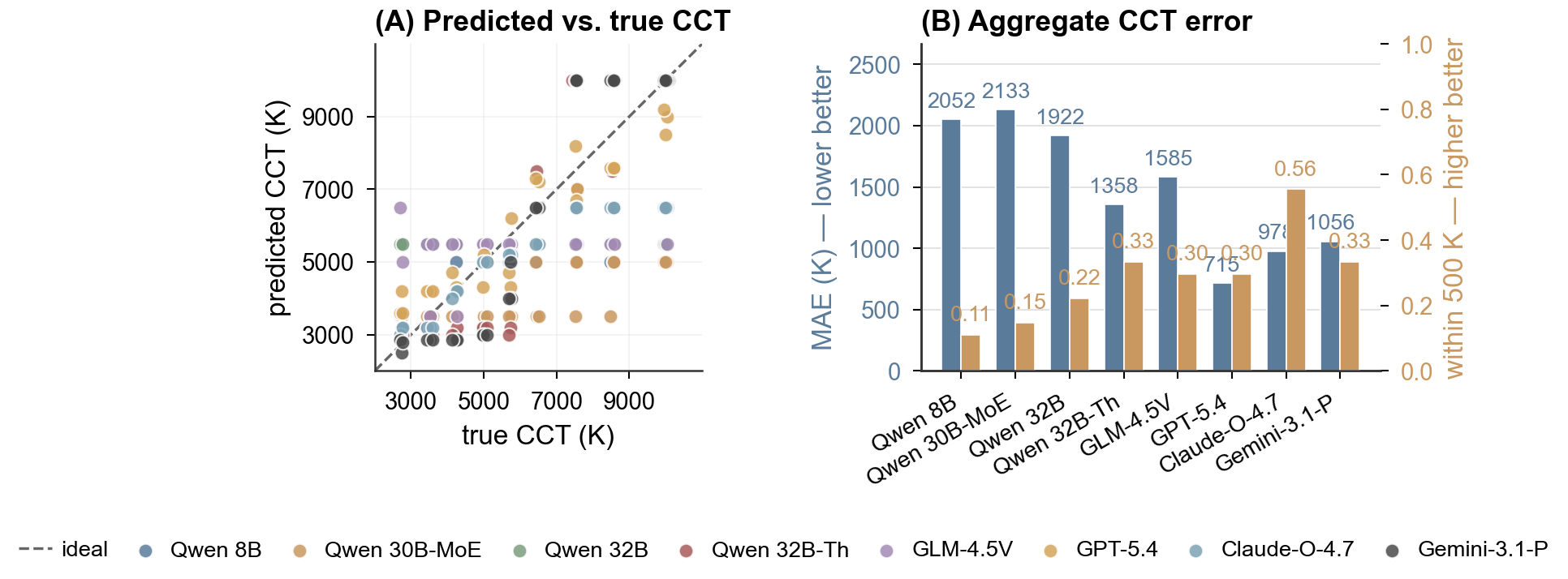}
\caption{\textbf{Task C --- CCT estimation across the
full 9-CCT, 27-stimulus task.}
(A)~Predicted vs.\ true CCT scatter. Five models (the four
Qwen variants + GLM-4.5V) cluster around a dashed $5500$\,K
daylight-band anchor (closer to D55 than to D65) regardless of
the true illuminant; GPT-5.4, Claude-Opus-4.7, and Gemini-3.1-Pro
track the identity line.
(B)~Aggregate mean absolute error (lower is better, blue, left
axis) and within-500\,K rate (higher is better, orange, right
axis). GPT-5.4 cuts MAE by $2.9\times$ over Qwen-8B;
Claude-Opus-4.7 reaches the highest within-500\,K rate of any
model.}
\label{fig:cct_calib}
\end{figure*}

\noindent\textbf{Failure mode 2: 5500\,K central-tendency anchor
(shared-prior cluster)}~Five of eight models return CCT values in
a narrow daylight band regardless of the true
illuminant (\figref{fig:cct_calib}A). The most extreme case is
Qwen-32B-Instruct, which returns 5500\,K for \emph{all 27 of 27}
stimuli --- a single-value predictor whose response is
independent of the input. The other four anchored models cluster
in the daylight band but with model-specific anchors (e.g.\
Qwen-30B-MoE collapses to 3500\,K and 5000\,K but never 5500\,K;
Qwen-8B is bimodal at 5500/5000\,K). The shared property is
therefore a daylight-band central-tendency regression, not a
literal 5500\,K collapse. Note that 5500\,K is closer to D55
($\sim$5503\,K) than to D65 (6504\,K) and lies below the K-space
midpoint ($\sim$6350\,K) and median (5700\,K) of our GT --- more
consistent with an sRGB-encoding prior than a D65-illuminant
prior. Mean absolute error (\figref{fig:cct_calib}B, blue bars)
splits into two clusters: 1358--2133\,K for the five anchored
models and 715--1056\,K for the three frontier closed-weight
models, a $2$--$3\times$ reduction (per-model values in
Table~\ref{tab:headline}).
Within-500\,K rate (orange bars) tells the same story: Claude
reaches 0.556 ($\sim 5\times$ the Qwen-8B baseline), and the only
anchored model that gets within striking distance is
Qwen-32B-Thinking at 0.333.

The cross-family test rules out a single-model-family effect:
GLM-4.5V, trained on an independent corpus and visual encoder,
also clusters its 27 predictions in the 5500--6500\,K band. But
the three frontier closed-weight models break the anchor: GPT-5.4
tracks identity within $\sim$700\,K on average, Claude-Opus-4.7
reaches the highest within-500\,K rate, and Gemini-3.1-Pro spreads
its predictions across the full 2700--10000\,K range. The shared anchor across Qwen and GLM points to a common prior
inside the open-weight cluster, but our experiments do not
identify the source: candidate explanations include over-
representation of daylight-balanced sRGB photography in the
open-weight pretraining ecosystem and reinforcement learning from
human feedback (RLHF) preference tuning that shapes numeric
outputs in the closed-weight models. Neither is
testable from the benchmark alone.

\begin{figure}[t]
\centering
\includegraphics[width=\linewidth]{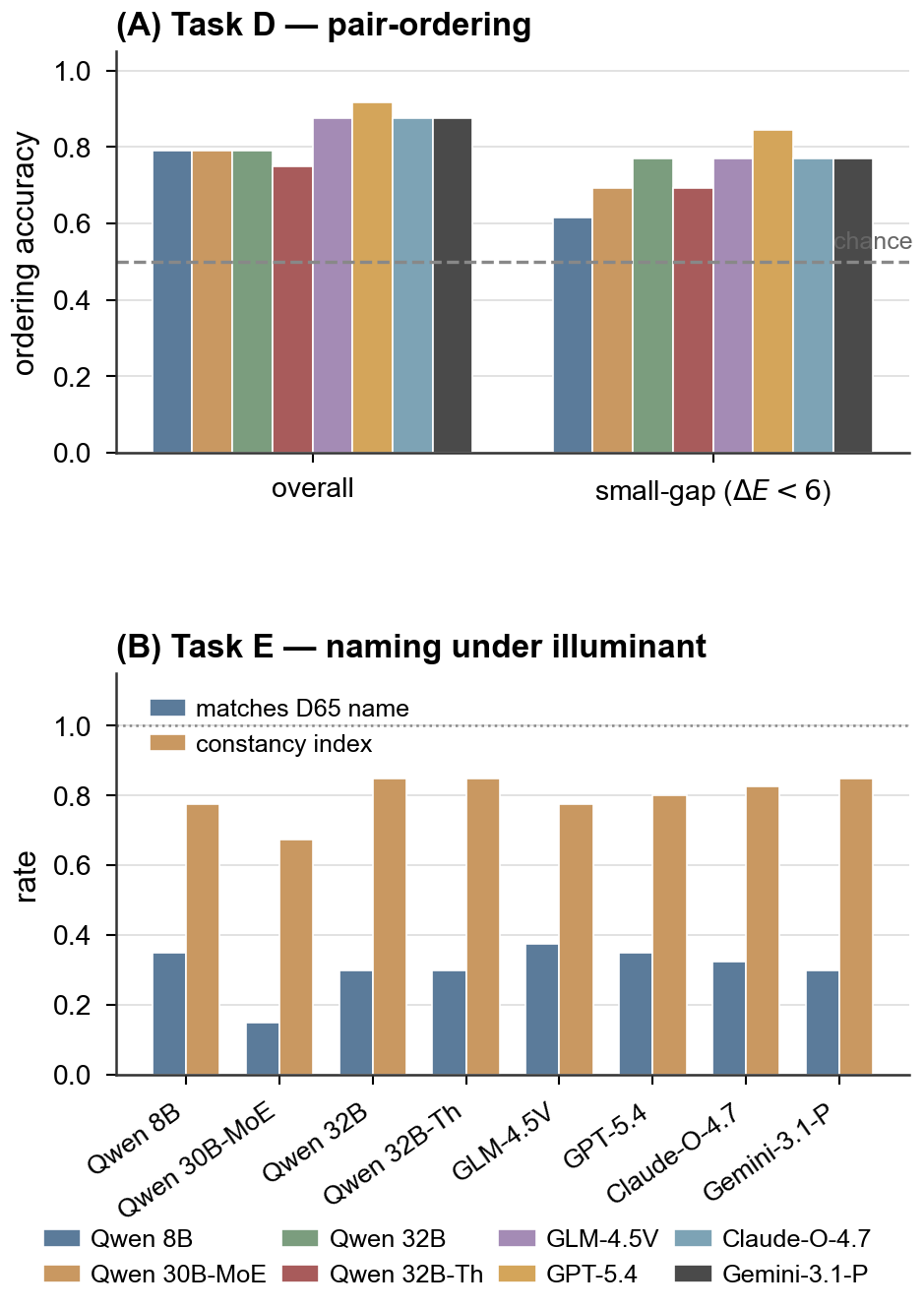}
\caption{\textbf{Tasks D and E, eight models.}
(A)~$\Delta E_{2000}$ pair-ordering accuracy (chance $=0.5$). The
four Qwen variants cluster in a tight $0.75$--$0.79$ plateau;
GLM-4.5V, Claude-Opus-4.7, Gemini-3.1-Pro all reach $0.875$ and
GPT-5.4 reaches $0.917$. The plateau is Qwen-specific.
(B)~Matches-D65-name accuracy (blue) and cross-illuminant
constancy index (orange) per model. Constancy clusters in a
$0.675$--$0.850$ band across all eight models; GLM-4.5V tops
matches-D65 at 0.375.}
\label{fig:taskDE}
\end{figure}

\noindent\textbf{Failure mode 3: $\Delta E$ ordering plateau
(Qwen-specific)}~All four Qwen variants plateau at $0.75$--$0.79$
$\Delta E_{2000}$ ordering accuracy; GLM-4.5V, Claude-Opus-4.7,
and Gemini-3.1-Pro reach $0.875$ and GPT-5.4 reaches $0.917$
(\figref{fig:taskDE}A). Among the 4 of 24 triplets where
linear-sRGB distance and $\Delta E_{2000}$ rank pairs differently,
no model consistently follows the RGB ranking. The plateau is
therefore better described as weak perceptual calibration on
small $\Delta E$ than as a simple RGB-distance rule.
\\

\noindent\textbf{Failure mode 4: D65-grounded naming bottleneck on
Task E}~The raw constancy index is misleading because a model
that locks onto \emph{any} stable name --- correct or not ---
scores high (\figref{fig:taskDE}B; Qwen-8B's 0.775 includes
``brown'' answered for nearly every warm-toned patch). Under
\emph{conditional constancy} --- consistency computed only on
patches whose 6500\,K baseline name matches D65 ground truth ---
the apparent $0.68$--$0.85$ spread collapses: 7 of 8 models tie
at $0.733$, only Qwen-30B-MoE drops to $0.500$, and the frontier
advantage disappears. Only 2--3 of 8 patches qualify per model
(yellow, purple, brown are mis-named at D65 by nearly every
model). The bottleneck is D65-grounded naming, not cross-illuminant
stability. Task B shows the same vocabulary bias on a different
stimulus: under the adjacent-clock chance of $0.375$ ($n{=}48$),
only GPT-5.4 ($0.79$) and Gemini-3.1-Pro ($0.75$) are
clearly above chance, and cool skin-tone shifts are
consistently mis-named as red or yellow.
\\

\noindent\textbf{Failure mode 5: Claude-Opus-4.7 pick-vs-name
asymmetry}~Claude names the cast direction perfectly on Task A
(bias $=1.000$) but picks the correctly-white-balanced image only
$0.667$ of the time. The 8 wrong-which trials split 4-A/4-B
(matching the ground-truth A/B distribution), ruling out a
position bias. We report this only as an observation to be tested
in future work; possible explanations include prompt effects and
sensitivity to the red--green axis near D65.
\\

\noindent\textbf{Robustness check: non-ColorChecker control}~As
ColorChecker-24 is heavily seen in training data, we re-rendered
Task C with 24 random sRGB-cube patches (identical pipeline
otherwise) and re-ran all 8 models. GPT-5.4
(715\,K\,$\to$\,1417\,K), Gemini-3.1-Pro (1056\,K\,$\to$\,1793\,K),
and Qwen-32B-Thinking (1358\,K\,$\to$\,2189\,K, the largest
degradation in the panel) all lose substantial accuracy without
the CC template, indicating that the frontier and chain-of-thought advantages
come partly from template-recognition priors.
Qwen-32B-Instruct holds at 1922\,K, confirming its daylight-band
anchor is a model property and not a template artifact.
Claude-Opus-4.7 is the only model that \emph{improves} on
template removal (978\,K\,$\to$\,752\,K).

\FloatBarrier

\section{Discussion}
\label{sec:discussion}

The eight-model set lets us group the findings by what changes
across models: model family, vendor access tier, or single-model
identity. We discuss practical use, failure grouping, and
limitations in turn.

\noindent\textbf{Use in color-aware pipelines}~Today's MLLMs
cannot replace classical camera-characterisation tools for fine
color judgments. That said, GPT-5.4 and Gemini-3.1-Pro both score 1.000 on both
halves of Task A. Flagship MLLMs can therefore serve as
\emph{coarse} screening checks in pipelines where no physical
ground truth is on hand. As order-of-magnitude calibration points: human chromatic
illumination-discrimination thresholds reported by Pearce et
al.~\cite{pearce2014} are on the order of $\sim$3--7 perceptual
$u'v'$ steps near a 6700\,K daylight reference; we do not project
this onto Kelvin because Kelvin is not perceptually uniform. On
the engineering side, classical learning-based AWB on
Gehler--Shi reaches mean angular error
$\sim\!1.78^{\circ}$~\cite{barron2017ffcc}, well below the
chromatic-discrimination range of any MLLM in our evaluation. The frontier MLLM CCT MAE of
715--1056\,K therefore falls in a range comparable in magnitude to
reported human illumination-discrimination thresholds near
daylight, while remaining far from specialised AWB systems. The open-weight cluster sits well above this floor on
CCT but matches frontier on binary AWB, so the pipeline choice
depends on whether the downstream task asks for a binary judgment
or a numeric Kelvin reading.
\\

\noindent\textbf{A failure-mode taxonomy}~The matrix sorts FM1--FM5
into four classes that a single-family study cannot tell apart:
\emph{Qwen-family-specific} (FM1, FM3 --- warm-bias and $\Delta
E$ plateau, source unidentified by the benchmark);
\emph{shared-prior cluster} (FM2 --- daylight-band anchor that
survives the Qwen$\to$GLM probe but breaks at the
open-weight$\to$frontier boundary); \emph{D65-grounded naming
bottleneck} (FM4 --- across all eight models, read as a possible
ceiling under this stimulus/prompt protocol, not architectural);
and \emph{model-specific} (FM5 Claude pick-vs-name, kept as one
hypothesis among several).
\\

\noindent\textbf{Limitations}~Stimuli are synthetic charts, not
natural scenes; a future version will use SFU AWB and Gehler--Shi
photographs. Three caveats: (a) \emph{pipeline confounds} ---
the metameric shortcut (sRGB values treated as reflectance under
D65) and the different visual-encoder tokenisation across the
eight models may both contribute to cross-model differences,
especially on Task A; v2 will use measured BabelColor spectra and
report tokenisation statistics. (b) \emph{prompt sensitivity} ---
a single fixed prompt per task cannot separate prompt effects
from chromatic perception; a 3-prompt ablation is deferred to v2.
(c) \emph{numeric output confound} --- CCT-in-Kelvin may test
calibrated numeric output as much as chromatic estimation; a
CCT-bin classification variant will accompany the regression in
v2. Without a human baseline we cannot claim MLLMs are
\emph{worse} than humans, only that they do not match
colorimetric ground truth; a 30-participant study on Tasks A and
B is underway.

\section{Conclusion}
\label{sec:conclusion}

We introduced a five-task, CIE-computable benchmark for
evaluating MLLM color judgment. Across eight endpoints, coarse AWB decisions
are often solved, but CCT estimation, $\Delta E_{2000}$ ordering,
and grounded color naming remain poorly calibrated. The strongest
split appears in Task C: tested open-weight endpoints show
daylight-band central-tendency anchors, while frontier endpoints
reduce CCT error --- partly through ColorChecker-template priors,
as the non-CC control reveals. The benchmark is limited to synthetic chart stimuli, a single
prompt per task, and sRGB-based rather than spectral rendering;
we plan to address these in the next benchmark version. 


\end{document}